%% file: main.tex
\pdfoutput=1
\documentclass{style}
\usepackage{array,makecell,tabularx}
\usepackage{algorithm}
\usepackage{algpseudocode}

\input{math_commands.tex}

\DeclareTextFontCommand{\textbf}{\bfseries}
\renewenvironment{figure*}[1][t]{\begin{figure}[#1]}{\end{figure}}
\input{sections/table_layout.tex}

\newcommand{\xt}{\mathbf{x}_t}

\newcommand{\epz}{\boldsymbol{\epsilon}}

\newcommand{\cf}{\mathbf{c}}
\newcommand{\tablebf}[1]{{\bfseries #1}}
\input{sections/tables.tex}
\arxivPairDecodeTables
\newcommand{\systemSpeedup}{}
\newcommand{\ditSpeedup}{}
\input{sections/numbers.tex}

\title{\includegraphics[height=10.8mm]{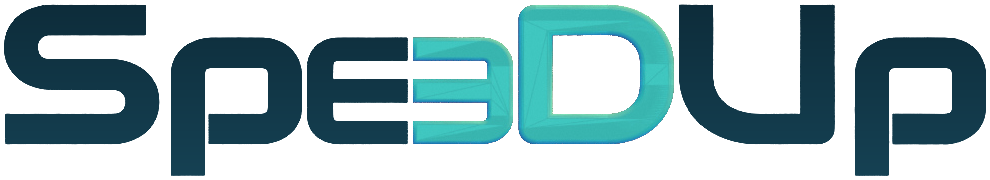}\\[0.28em]
Faithful Trajectory Distillation for Fast Image-to-3D Generation}
\author{%
  Liangchao~Xu\texorpdfstring{\textsuperscript{1,3,*}}{},\enspace
  Jiaye~Fu\texorpdfstring{\textsuperscript{2,3,*}}{},\enspace
  Youtian~Lin\texorpdfstring{\textsuperscript{1}}{},\enspace
  Feihu~Zhang\texorpdfstring{\textsuperscript{3}}{},\enspace
  Yao~Yao\texorpdfstring{\textsuperscript{1,\dag}}{}%
}
\affiliation[1]{Nanjing University}
\affiliation[2]{Peking University}
\affiliation[3]{Envision}
\contribution[*]{Equal contribution}
\contribution[\dag]{Corresponding author}
\hypersetup{
  pdftitle={Spe3DUp: Faithful Trajectory Distillation for Fast Image-to-3D Generation},
  pdfauthor={Liangchao Xu, Jiaye Fu, Youtian Lin, Feihu Zhang, Yao Yao}
}
\abstract{\input{sections/abstract.tex}}

\begin{document}
\maketitle
\begin{figure}[!htbp]
\centering
\includegraphics[width=\linewidth]{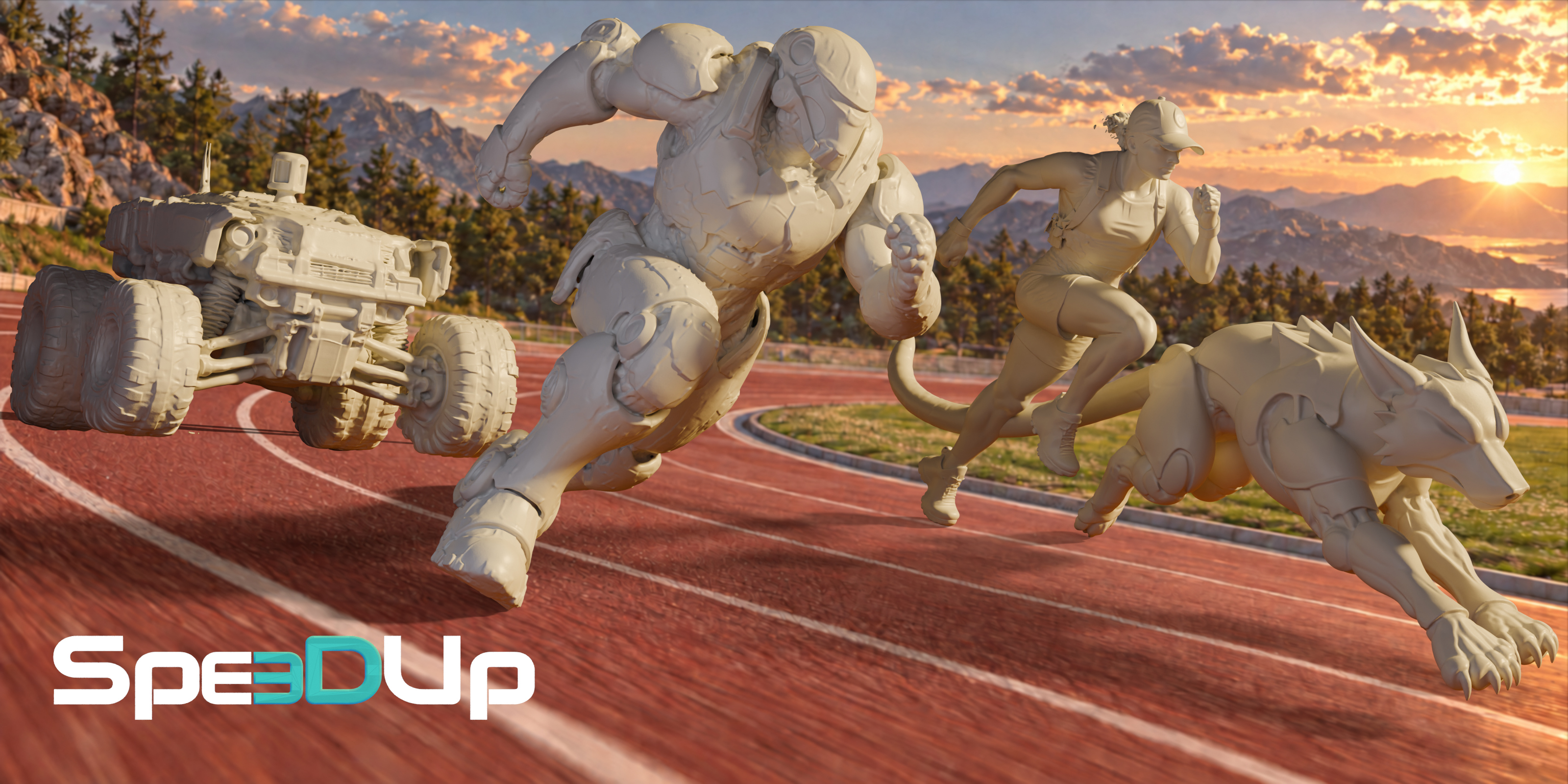}
\caption{\textbf{Spe3DUp: Faithful Trajectory Distillation for Fast Image-to-3D Generation.}}
\label{fig:teaser}
\end{figure}
\renderabstract
\vspace{2mm}
\input{sections/body.tex}
\FloatBarrier
{\small
\setlength{\bibsep}{2pt plus 0.5pt}
\bibliographystyle{ieeenat_fullname}
\bibliography{main}
}
\end{document}

%% file: math_commands.tex
\usepackage{amsmath,amsfonts,bm}

\def\eqref#1{equation~\ref{#1}}
\def\1{\bm{1}}

\def\vv{{\bm{v}}}

\DeclareMathAlphabet{\mathsfit}{\encodingdefault}{\sfdefault}{m}{sl}
\SetMathAlphabet{\mathsfit}{bold}{\encodingdefault}{\sfdefault}{bx}{n}

%% file: sections/table_layout.tex
\newcommand{\arxivtablefont}{\fontsize{10}{12}\selectfont}
\ExplSyntaxOn
\dim_new:N \l_arxiv_table_width_dim
\tl_new:N \l_arxiv_table_columns_tl
\tl_new:N \l_arxiv_table_body_tl
\bool_new:N \l_arxiv_table_inline_bool
\bool_new:N \l_arxiv_table_compact_bool
\cs_new_eq:NN \arxiv_table_float_begin: \table
\cs_new_eq:NN \arxiv_table_float_end: \endtable

\cs_new_protected:Npn \arxiv_table_layout:nnn #1#2#3
  {
    \exp_args:NV \str_if_in:nnT \l_arxiv_table_body_tl {\label{#1}}
      {
        \dim_set:Nn \l_arxiv_table_width_dim {#2\linewidth}
        \tl_set:Nn \l_arxiv_table_columns_tl {#3}
      }
  }
\cs_new_protected:Npn \arxiv_table_tabular:n #1
  {
    \begin{tabular*}{\l_arxiv_table_width_dim}
      {@{\extracolsep{\fill}}#1@{}}
  }
\NewDocumentEnvironment{arxivtabular}{m}
  {\exp_args:NV \arxiv_table_tabular:n \l_arxiv_table_columns_tl}
  {\end{tabular*}}
\cs_new_protected:Npn \arxiv_table_begin_star:n #1
  {\begin{arxivtabular}}
\cs_new_protected:Npn \arxiv_table_begin:n #1
  {
    \str_case:nnF {#1}
      {{tabular}{\begin{arxivtabular}}
       {tabular*}{\arxiv_table_begin_star:n}}
      {\begin{#1}}
  }
\cs_new_protected:Npn \arxiv_table_end:n #1
  {
    \str_case:nnF {#1}
      {{tabular}{\end{arxivtabular}}
       {tabular*}{\end{arxivtabular}}}
      {\end{#1}}
  }
\cs_new_protected:Npn \arxiv_table_setlength:nn #1#2
  {
    \tl_if_eq:nnTF {#1}{\tabcolsep}
      {\bool_if:NTF \l_arxiv_table_inline_bool
        {\setlength{\tabcolsep}{2.5pt}}
        {\bool_if:NTF \l_arxiv_table_compact_bool
          {\setlength{\tabcolsep}{2.5pt}}
          {\setlength{\tabcolsep}{7pt}}}}
      {\setlength{#1}{#2}}
  }

\cs_new_protected:Npn \arxiv_table_format:n #1
  {
    \tl_set:Nn \l_arxiv_table_body_tl {#1}
    \bool_set_false:N \l_arxiv_table_compact_bool
    \dim_set:Nn \l_arxiv_table_width_dim {\linewidth}
    \tl_clear:N \l_arxiv_table_columns_tl
    \arxiv_table_layout:nnn {tab:main} {0.90} {lcccccc}
    \arxiv_table_layout:nnn {tab:stage_consistency} {1.00} {lrrrr}
    \arxiv_table_layout:nnn {tab:fixed_coarse} {0.72} {lcccc}
    \arxiv_table_layout:nnn {tab:cross_cascade} {1.00} {llrrr}
    \arxiv_table_layout:nnn {tab:dmd_budget_gt100} {1.00} {lrrrr}
    \arxiv_table_layout:nnn {tab:rigid_input_alignment} {0.90} {lrrrr}
    \exp_args:NV \str_if_in:nnT \l_arxiv_table_body_tl {\label{tab:stage_consistency}}
      {\bool_set_true:N \l_arxiv_table_compact_bool}
    \arxiv_table_layout:nnn {tab:cumulative_acceleration} {0.86} {lrrr}
    \arxiv_table_layout:nnn {tab:decode_efficiency} {0.82} {lrr}
    \arxiv_table_layout:nnn {tab:refinement_decode_gt} {0.72} {lccccc}
    \arxiv_table_layout:nnn {tab:refinement_decode_quality} {0.86} {lrrr}
    \arxiv_table_layout:nnn {tab:coarse_geometry} {1.00} {lcrrrrr}
    \arxiv_table_layout:nnn {tab:training_compute} {0.94} {lrrr}
    \arxiv_table_layout:nnn {tab:current_training_config} {1.00} {lrrrrlll}
    \arxiv_table_layout:nnn {tab:training_evaluation_grids} {0.94} {lcc}
    \arxiv_table_layout:nnn {tab:paired_geometry75_uncertainty} {1.00} {lllrlll}
    \arxiv_table_layout:nnn {tab:paired_rigid74_uncertainty} {1.00} {lllrlll}
    \arxiv_table_layout:nnn {tab:paired_image60_uncertainty} {1.00} {lllrlll}
    \arxiv_table_layout:nnn {tab:gt100_coarse_geometry} {1.00} {lrrrrrr}
    \arxiv_table_layout:nnn {tab:gt100_components} {1.00} {lrrrrrrr}
    \arxiv_table_layout:nnn {tab:gt100_paired_uncertainty} {1.00} {lrrrr}
    \arxiv_table_layout:nnn {tab:refinement_decode_topology_audit} {1.00} {lr}
    \clist_map_inline:nn
      {tab:current_training_config,tab:paired_geometry75_uncertainty,
       tab:paired_rigid74_uncertainty,tab:paired_image60_uncertainty}
      {
        \exp_args:NV \str_if_in:nnT \l_arxiv_table_body_tl {\label{##1}}
          {
            \bool_set_true:N \l_arxiv_table_compact_bool
            \renewcommand{\arxivtablefont}{\fontsize{9}{11}\selectfont}
            \tl_replace_all:Nnn \l_arxiv_table_body_tl
              {Paired/attempted} {\makecell{Paired/\\attempted}}
            \tl_replace_all:Nnn \l_arxiv_table_body_tl
              {Mean~difference} {\makecell{Mean\\difference}}
            \tl_replace_all:Nnn \l_arxiv_table_body_tl
              {Time~starts~/~sampling} {\makecell{Time~starts/\\sampling}}
          }
      }
    \arxiv_table_layout:nnn {tab:dmd_continuation} {0.94} {lrrrr}
    \arxiv_table_layout:nnn {tab:dmd_final_geometry} {1.00} {lrrrrrrr}
    \arxiv_table_layout:nnn {tab:dmd_final_consistency} {0.94} {lrrrr}
    \arxiv_table_layout:nnn {tab:system_efficiency} {0.94} {lrrrr}
    \arxiv_table_layout:nnn {tab:decode_ablation} {0.76} {lrrr}
    \arxiv_table_layout:nnn {tab:decoder_four_path} {0.58} {lcc}
    \arxiv_table_layout:nnn {tab:image100_primary_native} {0.94} {lrrrr}
    \arxiv_table_layout:nnn {tab:steps} {0.86} {ccrrrr}
    \arxiv_table_layout:nnn {tab:remesh} {0.92} {rrrrrr}
    \arxiv_table_layout:nnn {tab:remesh_equiv} {0.82} {lrrr}
    \bool_if:NT \l_arxiv_table_inline_bool
      {\dim_set:Nn \l_arxiv_table_width_dim {\linewidth}}
    \tl_if_empty:NT \l_arxiv_table_columns_tl
      {\PackageError{arxiv-tables}{Missing~single-column~table~layout}
        {Add~the~new~table~label~to~arxiv/table_layout.tex.}}
    \captionsetup{width=\l_arxiv_table_width_dim,
      justification=raggedright,singlelinecheck=false,skip=6pt}
    
    \tl_replace_all:Nnn \l_arxiv_table_body_tl {\small} {\arxivtablefont}
    \tl_replace_all:Nnn \l_arxiv_table_body_tl
      {\setlength} {\arxiv_table_setlength:nn}
    \tl_replace_all:Nnn \l_arxiv_table_body_tl
      {\begin} {\arxiv_table_begin:n}
    \tl_replace_all:Nnn \l_arxiv_table_body_tl
      {\end} {\arxiv_table_end:n}
    \setlength{\tabcolsep}{7pt}
    \tl_use:N \l_arxiv_table_body_tl
  }
\RenewDocumentEnvironment{table}{O{t}+b}
  {\bool_if:NTF \l_arxiv_table_inline_bool
    {\arxiv_table_format:n {#2}}
    {\arxiv_table_float_begin:[t]
      \arxiv_table_format:n {#2}\arxiv_table_float_end:}}
  {}
\RenewDocumentEnvironment{table*}{O{t}+b}
  {\bool_if:NTF \l_arxiv_table_inline_bool
    {\arxiv_table_format:n {#2}}
    {\arxiv_table_float_begin:[t]
      \arxiv_table_format:n {#2}\arxiv_table_float_end:}}
  {}
\NewDocumentCommand{\arxivPairEfficiencyTables}{}
  {
    \cs_set_eq:NN \arxiv_cumulative_source: \cumulativeaccelerationtable
    \cs_set_eq:NN \arxiv_decode_source: \decodeefficiencytable
    \RenewDocumentCommand{\cumulativeaccelerationtable}{}
      {
        \arxiv_table_float_begin:[!htbp]
        \centering
        \bool_set_true:N \l_arxiv_table_inline_bool
        \renewcommand{\arxivtablefont}{\fontsize{9}{11}\selectfont}
        \begin{minipage}[t]{0.51\linewidth}
          \arxiv_cumulative_source:
        \end{minipage}\hfill
        \begin{minipage}[t]{0.46\linewidth}
          \arxiv_decode_source:
        \end{minipage}
        \arxiv_table_float_end:
      }
    \RenewDocumentCommand{\decodeefficiencytable}{}{}
  }
\NewDocumentCommand{\arxivPairDecodeTables}{}
  {
    \cs_set_eq:NN \arxiv_decode_gt_source: \refinementdecodegttable
    \cs_set_eq:NN \arxiv_decode_prof_source: \decoderprofilestable
    \RenewDocumentCommand{\refinementdecodegttable}{}
      {
        \arxiv_table_float_begin:[t]
        \centering
        \bool_set_true:N \l_arxiv_table_inline_bool
        \renewcommand{\arxivtablefont}{\fontsize{9}{11}\selectfont}
        \begin{minipage}[t]{0.50\linewidth}
          \vspace{0pt}
          \arxiv_decode_gt_source:
        \end{minipage}\hfill
        \begin{minipage}[t]{0.46\linewidth}
          \vspace{0pt}
          \arxiv_decode_prof_source:
        \end{minipage}
        \arxiv_table_float_end:
      }
    \RenewDocumentCommand{\decoderprofilestable}{}{}
  }

\ExplSyntaxOff

%% file: sections/tables.tex
\newcommand{\primarygeometrytable}{%
\begin{table*}[t]
\caption{\textbf{Geometry and input fidelity of coarse generation and complete cascades.}
\mbox{G: geometry}; \mbox{I: input consistency}.
CD is multiplied by $10^3$; best few-step results are bold.}
\label{tab:main}
\centering\small
\setlength{\tabcolsep}{3.5pt}
\begin{tabular*}{\linewidth}{@{\extracolsep{\fill}}lcccccc@{}}
\toprule
Method & Steps & CD$^{\mathrm{G}}\downarrow$ & F@.01$^{\mathrm{G}}\uparrow$ & Normal$^{\mathrm{G}}\uparrow$ & IoU$^{\mathrm{I}}\uparrow$ & ULIP-2$^{\mathrm{I}}\uparrow$ \\
\midrule
\multicolumn{7}{@{}l}{\textit{Coarse generation}} \\
Teacher & 50 & 15.3 & 0.596 & 0.751 & 0.844 & 0.237 \\
DMD & 10 & 28.7 & 0.419 & 0.718 & 0.729 & 0.210 \\
Spe3DUp & 10 & \tablebf{15.4} & \tablebf{0.589} & \tablebf{0.752} & \tablebf{0.840} & \tablebf{0.232} \\
\midrule
\multicolumn{7}{@{}l}{\textit{Complete cascade (coarse/refinement steps)}} \\
Teacher & 50/50 & 15.7 & 0.575 & 0.737 & 0.839 & 0.233 \\
DMD & 10/5 & 29.1 & 0.413 & 0.706 & 0.725 & 0.195 \\
Spe3DUp & 10/5 & \tablebf{15.2} & \tablebf{0.583} & \tablebf{0.742} & \tablebf{0.837} & \tablebf{0.232} \\
\bottomrule
\end{tabular*}
\end{table*}
}

\newcommand{\fixedcoarserefinementtable}{%
\begin{table}[!t]
\caption{\textbf{Refinement with fixed teacher coarse geometry.}
\mbox{G: geometry}; \mbox{I: input consistency}.
CD is multiplied by $10^3$; best few-step results are bold.}
\label{tab:fixed_coarse}
\centering\small
\setlength{\tabcolsep}{3pt}
\begin{tabular*}{\columnwidth}{@{\extracolsep{\fill}}lrrrr@{}}
\toprule
Refiner & Steps & CD$^{\mathrm{G}}\downarrow$ & F@.01$^{\mathrm{G}}\uparrow$ & ULIP-2$^{\mathrm{I}}\uparrow$ \\
\midrule
Teacher & 50 & 15.7 & 0.575 & 0.233 \\
DMD & 5 & 15.9 & 0.572 & 0.213 \\
Spe3DUp & 5 & \tablebf{15.4} & \tablebf{0.580} & \tablebf{0.235} \\
\bottomrule
\end{tabular*}
\end{table}
}

\newcommand{\refinementdecodegttable}{%
\begin{table}[!t]
\caption{\textbf{Geometric quality of original and accelerated decoding.}}
\label{tab:refinement_decode_gt}
\centering\small
\setlength{\tabcolsep}{2pt}
\begin{tabular*}{\columnwidth}{@{\extracolsep{\fill}}lrrrrr@{}}
\toprule
Decoder & CD $\downarrow$ & F@.01 $\uparrow$ & F@.02 $\uparrow$ & NC $\uparrow$ & HD95 $\downarrow$ \\
\midrule
Original & 0.0356 & 0.2621 & 0.4510 & 0.5835 & 0.1140 \\
Accelerated & 0.0356 & 0.2620 & 0.4510 & 0.5835 & 0.1140 \\
\bottomrule
\end{tabular*}
\end{table}
}

\newcommand{\decoderprofilestable}{%
\begin{table}[!t]
\caption{\textbf{Refinement decoder configurations.}
Field-evaluation time is measured on A800; surface distance is the mean symmetric deviation from the reference mesh.}
\label{tab:decoder_four_path}
\centering\small
\setlength{\tabcolsep}{3pt}
\begin{tabular*}{\columnwidth}{@{\extracolsep{\fill}}lrr@{}}
\toprule
Configuration & Field (s) $\downarrow$ & Surface dist. ($\times 10^{-4}$) $\downarrow$ \\
\midrule
Reference & 25.6 & 0.00 \\
Operator-only & 14.1 & 0.55 \\
Field reuse & 11.1 & 6.59 \\
Recomputation & 12.3 & 1.52 \\
\bottomrule
\end{tabular*}
\end{table}
}

%% file: sections/numbers.tex
\providecommand{\teacherBaseTime}{576.6}

\providecommand{\studentEfficientTime}{74.5}
\providecommand{\teacherEfficientTime}{437.7}

\renewcommand{\systemSpeedup}{5.87}
\renewcommand{\ditSpeedup}{18.88}
\providecommand{\cumulativeSpeedup}{7.74}

\providecommand{\coarseDitSpeedup}{10.0}
\providecommand{\refineDitSpeedup}{19.9}
\providecommand{\cumulativeSpeedupRounded}{7.7}

%% file: sections/abstract.tex
Recent 3D generative models can produce high-quality meshes with complex structures and fine geometric details, with broad applications in games, film, and digital design.
However, high-quality generation typically relies on long multi-step sampling processes that repeatedly evaluate large networks, leading to long inference times.
At the same time, reference-driven tasks such as game-asset reconstruction and product digitization demand close correspondence to the input image: generated geometry must faithfully reconstruct the specific reference object.
Acceleration must therefore reduce sampling computation while preserving the model's ability to reconstruct that object.
To enable fast and faithful 3D generation, we introduce \textbf{Spe3DUp}, a coarse-to-fine framework based on \mbox{\textbf{Faithful Trajectory Distillation (FTD)}}.
FTD directly supervises few-step students with the teacher's input-conditioned denoising trajectories to preserve its generation behavior.
We apply FTD separately to coarse geometry generation and geometric refinement, compressing both sampling processes while maintaining correspondence to the reference input.
We further accelerate the complete mesh-generation pipeline with efficient geometry decoding and parallel remeshing.
Experiments show that Spe3DUp maintains geometric quality and input correspondence close to those of the teacher, accelerates the coarse and refinement DiTs by $\coarseDitSpeedup\times$ and $\refineDitSpeedup\times$, respectively, and achieves a $\cumulativeSpeedupRounded\times$ end-to-end speedup over the original teacher pipeline.

%% file: sections/body.tex
\section{Introduction}

Recent 3D generative models can produce meshes with complex structures and fine surface details~\citep{wu2024direct3d,zhang2024clay,xiang2024trellis,hunyuan3d2025}.
Progress in global shape modeling and local geometry recovery brings automated asset creation closer to practical use, making generation efficiency increasingly important~\citep{ultrashape2025}.

However, applications such as recreating game characters and props, digitizing products, and reconstructing existing designs demand more than detailed surfaces.
Generated meshes must faithfully reproduce the object specified by the input image, including its shape, proportions, and part layout.
A visually plausible model with rich detail may still be unsuitable if it changes a character's body proportions, the placement of mechanical parts, or a product's overall silhouette.

High-quality 3D generation typically relies on multi-step diffusion sampling, repeatedly evaluating large networks at substantial computational cost.
Guidance distillation, progressive distillation, and distribution matching offer different routes to reducing this cost~\citep{meng2023distillation,salimans2022progressive,yin2024dmd,yin2024dmd2}.
We aim to accelerate coarse-to-fine 3D generation while preserving the teacher's geometric response to the input image.

We present \textbf{Spe3DUp}, centered on \textbf{Faithful Trajectory Distillation (FTD)} (\cref{fig:overview}).
FTD builds on guidance distillation and multi-step-to-one-step endpoint regression~\citep{meng2023distillation,salimans2022progressive}.
FTD trains coarse and refinement students to preserve the teacher's response to each reference input.

After sampling is compressed, latent decoding, surface extraction, and mesh processing account for a larger fraction of total latency and limit further acceleration.
We use hierarchical field queries for coarse geometry and design cross-resolution scalar-field reuse with critical-region recomputation for high-resolution refinement decoding.
This reduces repeated neural-field queries while concentrating additional evaluations in geometrically sensitive regions.
Region-parallel mesh simplification further reduces geometry-processing overhead, translating sampling gains into faster mesh delivery.

We evaluate Spe3DUp on a generation pipeline combining Hunyuan3D-2.1 and UltraShape.
We examine coarse generation, refinement under fixed coarse geometry, and complete cascade outputs.
Experiments show that FTD maintains average geometric quality and input correspondence close to those of the teacher while accelerating diffusion sampling by $\ditSpeedup\times$.
With optimized decoding and postprocessing shared by teacher and student, end-to-end latency decreases from $\teacherEfficientTime$ to $\studentEfficientTime$ seconds ($\systemSpeedup\times$); relative to the original unoptimized teacher ($\teacherBaseTime$ seconds), the cumulative system gain is $\cumulativeSpeedup\times$.

Our contributions are twofold:
\begin{itemize}
\item We apply guidance and interval-endpoint distillation to coarse-to-fine 3D generation as \textbf{Faithful Trajectory Distillation}, using paired trajectory supervision to compress coarse generation and refinement while preserving the teacher's input-conditioned geometric response.
\item We design a high-resolution geometry-recovery acceleration strategy.
Cross-resolution scalar-field reuse and critical-region recomputation reduce redundant decoding queries, while parallel mesh simplification enables end-to-end mesh-generation acceleration.
\end{itemize}

\begin{figure*}[t]
\centering
\includegraphics[width=\linewidth]{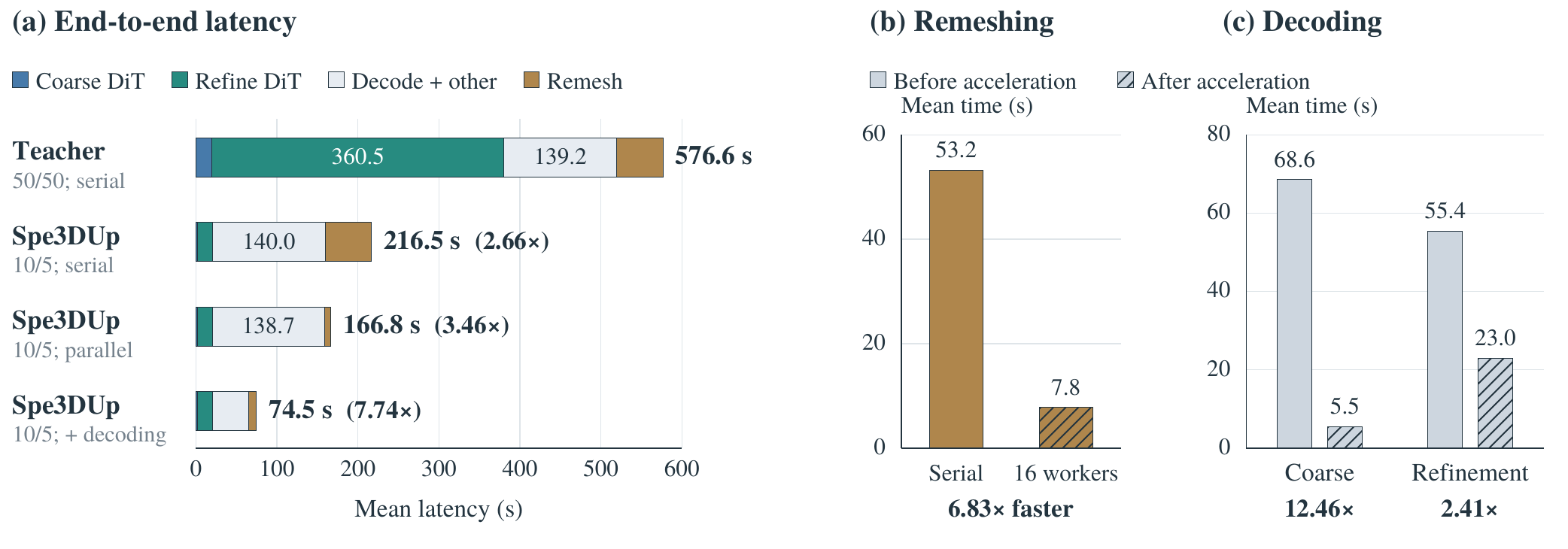}
\caption{\textbf{System acceleration.}
(a) End-to-end latency breakdown.
(b) Mesh simplification time.
(c) Coarse and refinement decoding time.}
\label{fig:latency_breakdown}
\end{figure*}

\section{Related Work}

\paragraph{Native 3D Generation.}
Native 3D generators synthesize complete geometry by modeling shape distributions in learned 3D latent spaces.
3DShape2VecSet introduces VecSet, a compact latent-vector set suited to transformer-based shape generation~\citep{zhang2023vecset}.
Direct3D, CLAY, and Hunyuan3D build image-conditioned generative models in 3D latent spaces to synthesize complex objects~\citep{wu2024direct3d,zhang2024clay,hunyuan3d2025}.
Work on higher-resolution geometry synthesis explores both spatial organization and coarse-to-fine modeling.
TRELLIS models structured latents on sparse spatial supports~\citep{xiang2024trellis}.
LATTICE anchors latent vectors to a coarse voxel grid to form VoxSet, generating geometric anchors before synthesizing detailed geometry~\citep{lai2025lattice}.
Coarse-to-fine methods use an initial shape to guide subsequent detail recovery.
CraftsMan3D combines generation with a geometry refiner~\citep{li2024craftsman}, DetailGen3D learns coarse-to-fine transformations through data-dependent flows~\citep{deng2024detailgen3d}, and UltraShape conditions local detail synthesis on voxel queries derived from coarse geometry~\citep{ultrashape2025}.
Seed3D 2.0 and AssetGen also adopt coarse-to-fine geometry generation within asset pipelines~\citep{gu2026seed3d2,wang2026assetgen}.
Within this coarse-to-fine formulation, we distill pretrained generators to reduce the sampling cost of both geometry stages while preserving the teacher's input-conditioned reconstruction behavior.

\paragraph{Distillation for Efficient Generation.}
Progressive distillation reduces sampling steps by training one student update to match multiple teacher updates~\citep{salimans2022progressive}.
Meng et al.~\citep{meng2023distillation} combine guidance distillation with progressive distillation, first folding classifier-free guidance into a single conditional model and then compressing the sampling process.
Consistency models and their latent-space extension LCM learn mappings from different states on the same trajectory to a common endpoint~\citep{song2023consistency,luo2023lcm}.
InstaFlow uses reflow to straighten trajectories before distillation, facilitating few-step generation~\citep{liu2024instaflow}.
DMD combines distribution matching with paired teacher-output regression, whereas DMD2 removes that regression term and adds adversarial supervision~\citep{yin2024dmd,yin2024dmd2}.
ADD and LADD likewise use adversarial learning to improve few-step generation quality~\citep{sauer2023add,sauer2024ladd}.
In 3D generation, FlashVDM and MDT-dist explore distillation to reduce geometry-generation sampling costs~\citep{flashvdm2025,mdtdist2025}.
Following this direction, we study trajectory compression for coarse-to-fine generation, focusing on whether accelerated models retain the teacher's geometric response to specific inputs.
FTD uses paired teacher-trajectory supervision to compress coarse geometry generation and geometric refinement separately.

\section{Method}
\label{sec:method}

\subsection{Overview}
\label{sec:method_overview}

\begin{figure*}[t]
\centering
\includegraphics[width=\linewidth]{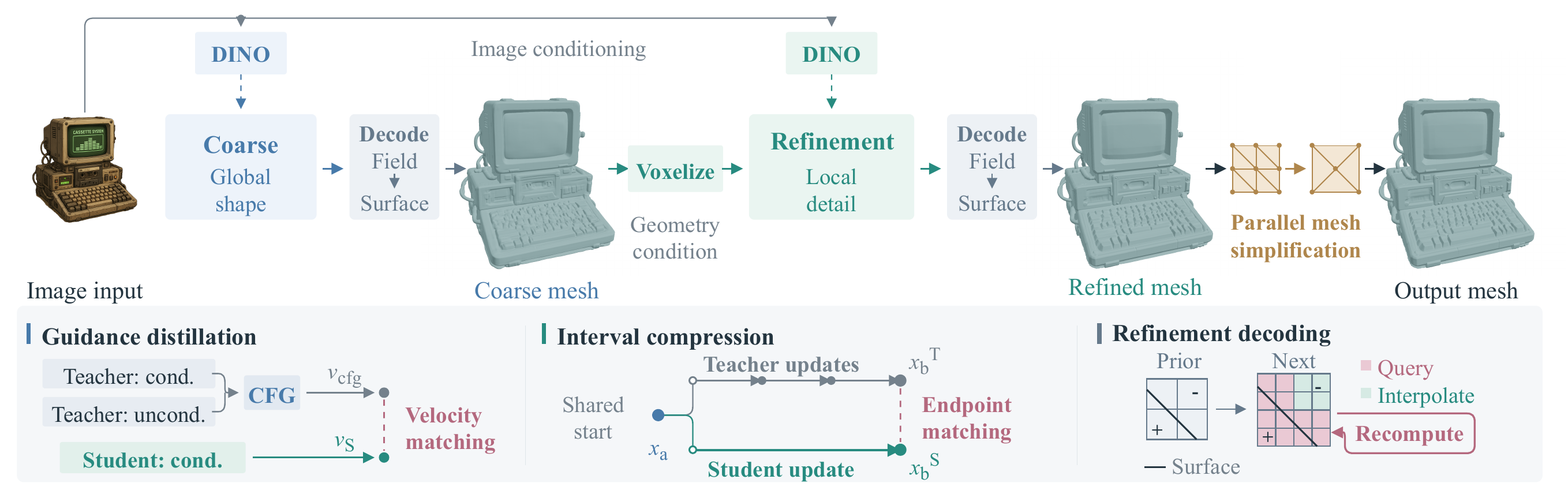}
\caption{\textbf{Overview of Spe3DUp.}
Top: the coarse-to-fine generation pipeline.
Bottom: FTD and decoding acceleration.}
\label{fig:overview}
\end{figure*}

Spe3DUp accelerates a two-stage pipeline comprising coarse geometry generation and geometric refinement (\cref{fig:overview}).
Given an image $I$, the coarse generator produces a mesh $\mathcal M^c$ describing the object's global structure.
The refiner then combines the image and coarse geometry to recover local surface detail, producing $\mathcal M^r$.
We index these stages by $k\in\{c,r\}$ and denote their conditions by
\begin{equation}
\cf^c=I,\qquad \cf^r=(I,\mathcal M^c).
\label{eq:stage_conditions}
\end{equation}

Both generators perform conditional flow sampling in a learned geometry latent space~\citep{lipman2023flow}.
For clean latent $\mathbf z^k$ and Gaussian noise $\epz$, we use a noise-to-data time convention,
\begin{equation}
\xt^k=(1-t)\epz+t\mathbf z^k,\qquad t\in[0,1].
\label{eq:latent_path}
\end{equation}
The model predicts a conditional velocity field, which a numerical solver integrates to obtain the latent for geometry decoding.
We use Faithful Trajectory Distillation (FTD) to compress this sampling process while retaining the teacher's geometric response to the reference image and, for refinement, the coarse geometry.
We first transfer classifier-free guidance into a single conditional branch, then train one student update to match multiple teacher updates.
Efficient geometry decoding and parallel mesh simplification reduce the remaining cost of delivering the final mesh.

\subsection{Faithful Trajectory Distillation}
\label{sec:cfg_distill}

\paragraph{Guidance distillation.}
We follow guidance distillation~\citep{meng2023distillation} to replace the two prediction branches of CFG~\citep{ho2022classifierfree} with one conditional model.
Let $\cf_{\mathrm{null}}^c=\varnothing_I$ and $\cf_{\mathrm{null}}^r=(\varnothing_I,\mathcal M^c)$, where $\varnothing_I$ denotes the stage's null image embedding.
The refinement null branch retains the same coarse-geometry conditioning, implemented as voxel anchors.
Given noisy latent $\xt^k$, the original teacher combines image-conditional and image-null velocities as
\begin{equation}
\vv_\phi^{k,\mathrm{cfg}}(\xt^k,t,\cf^k)
=\vv_\phi^k(\xt^k,t,\cf_{\mathrm{null}}^k)
+w^k\bigl(\vv_\phi^k(\xt^k,t,\cf^k)-\vv_\phi^k(\xt^k,t,\cf_{\mathrm{null}}^k)\bigr).
\label{eq:cfg}
\end{equation}
Each sampling step therefore evaluates both branches.
We initialize a conditional student from the teacher and regress the guided prediction using mean squared error:
\begin{equation}
\mathcal{L}_{\mathrm{cfg}}^k
=
\mathbb{E}
\left[
\lambda^k(t)
\left\|
\vv_\psi^k(\xt^k,t,\cf^k)
-
\vv_\phi^{k,\mathrm{cfg}}(\xt^k,t,\cf^k)
\right\|_2^2
\right].
\label{eq:l_cfg}
\end{equation}
Here $\lambda^c(t)=1$ for velocity regression and $\lambda^r(t)=(1-t)^2$, corresponding to the refinement implementation's regression of clean-latent predictions $\hat{\mathbf z}=\xt+(1-t)\vv$.
The teacher remains frozen during this optimization.
Once trained, the conditional model approximates the original guided prediction with one forward evaluation per sampling step.
We freeze its parameters, denoted $\bar\psi^k$, and use it as the teacher for interval distillation.

\paragraph{Interval compression.}
Following the multi-step-to-one-step principle of progressive distillation~\citep{salimans2022progressive}, we train one student update to approximate the teacher's cumulative update over the same interval.
For a selected interval $[a,b]$, we construct $\mathbf x_a^k$ from a data latent and noise using \cref{eq:latent_path}.
The appendix provides details of interval selection, teacher substeps, and the stage-specific training and inference grids.
Starting at $\mathbf x_0^{k,T}=\mathbf x_a^k$, the frozen teacher takes Euler substeps on $a=\tau_0<\cdots<\tau_m=b$:
\begin{equation}
\mathbf x_{\ell+1}^{k,T}
=\mathbf x_\ell^{k,T}+(\tau_{\ell+1}-\tau_\ell)\,\vv_{\bar\psi}^k(\mathbf x_\ell^{k,T},\tau_\ell,\cf^k).
\label{eq:teacher_substeps}
\end{equation}
We initialize the student from the guidance-distilled model.
With the same starting latent and condition, it crosses the entire interval in one forward evaluation:
\begin{equation}
\mathbf x_b^{k,S}
=\mathbf x_a^k+(b-a)\vv_\theta^k(\mathbf x_a^k,a,\cf^k).
\label{eq:flow_maps}
\end{equation}
The teacher endpoint directly supervises this update through mean squared error:
\begin{equation}
\mathcal{L}_{\mathrm{traj}}^k
=
\mathbb{E}\left[
\left\|
\mathbf{x}_{b}^{k,S}
-
\operatorname{sg}\!\left(\mathbf{x}_{m}^{k,T}\right)
\right\|_2^2
\right],
\label{eq:l_traj}
\end{equation}
where $\operatorname{sg}$ stops gradients through the teacher.
The endpoint loss trains the student to reproduce the teacher's accumulated conditional update over the interval, with both models sharing the same starting state and conditions.
For refinement, both the image and coarse geometry are identical in each teacher--student pair.
This directly supervises the student's approximation of the teacher's input-dependent sampling response.
We distill the two generators separately and cascade them at inference, with refinement conditioned on both the image and the generated coarse geometry.

\subsection{Efficient geometry decoding}
\label{sec:efficient_decode}

After FTD compresses sampling, latent decoding accounts for a larger fraction of inference.
The decoder encodes spatial positions as queries and geometry-latent features as keys and values.
Cross-attention aggregates these features, and a prediction head produces the scalar field for zero-isosurface extraction.
We use FlashVDM's hierarchical queries for coarse geometry~\citep{flashvdm2025} and reduce high-resolution refinement queries through field reuse and critical-region recomputation.

\paragraph{Cross-resolution field reuse.}
For a fixed latent, the preceding-resolution field estimates where the surface lies.
We query the neural decoder near potential surface cells and interpolate elsewhere, concentrating high-resolution evaluations on regions that may affect geometry.

\paragraph{Critical-region recomputation.}
Interpolation can change local signs and connectivity.
We recompute original query batches containing near-surface inconsistencies, reusing their previously selected attention keys and values.
Remaining sign-changing edge endpoints and near-surface queries are then evaluated before mesh extraction.
Algorithm~\ref{alg:refinement_decode} gives the field updates and stopping conditions; detection criteria are provided in the appendix.
\input{sections/algorithm.tex}

\subsection{Parallel mesh simplification}
\label{sec:parallel_remesh}

High-resolution surface extraction often produces far more triangles than the delivery budget.
Serial quadric-error-metric (QEM) simplification~\citep{garland1997surface} then requires many successive edge collapses.
Following spatially parallel simplification ~\citep{xiong2008parallel,decoro2007realtime,mousa2021highperformance}, we distribute most of this work across independent regions.

We first partition the mesh into spatially contiguous regions and allocate face budgets in proportion to their input face counts.
The combined regional face budget exceeds the final target, leaving a surplus for the subsequent global simplification pass.
Each region then runs QEM independently with protected partition boundaries, allowing most edge collapses to proceed in parallel.

Finally, we stitch the regional outputs and perform one global QEM pass to reach the target face budget.
The regional stage handles most face reduction, while the global stage resolves the remaining simplification and budget allocation on a smaller mesh.

\section{Experiments}
\label{sec:exp}

\subsection{Experimental setup}
\label{sec:exp_setup}

We evaluate FTD on a pipeline combining Hunyuan3D-2.1 for coarse geometry generation and UltraShape for refinement~\citep{hunyuan3d2025,ultrashape2025}.
The main comparison includes the original Teacher pipeline, DMD, and Spe3DUp with FTD applied to both generators.
Teacher uses 50 sampling steps with CFG at each stage; DMD and FTD use 10 coarse-generation steps and 5 refinement steps without runtime CFG.

We report Chamfer distance (CD), F-score at distance 0.01, and normal consistency for geometric reconstruction, together with silhouette IoU and ULIP-2~\cite{xue2024ulip2} for correspondence to the input image.
We first compare coarse generation and complete cascades, then isolate refinement by supplying all refiners with the same teacher coarse geometry.
Inference latency is measured on an NVIDIA L20 GPU.
The appendix provides dataset and training details, additional metrics, and cross-cascade combinations.

\subsection{Generation quality}
\label{sec:main_results}
\label{sec:stage_consistency}

\primarygeometrytable

\begin{figure*}[t]
\centering
\includegraphics[width=\linewidth]{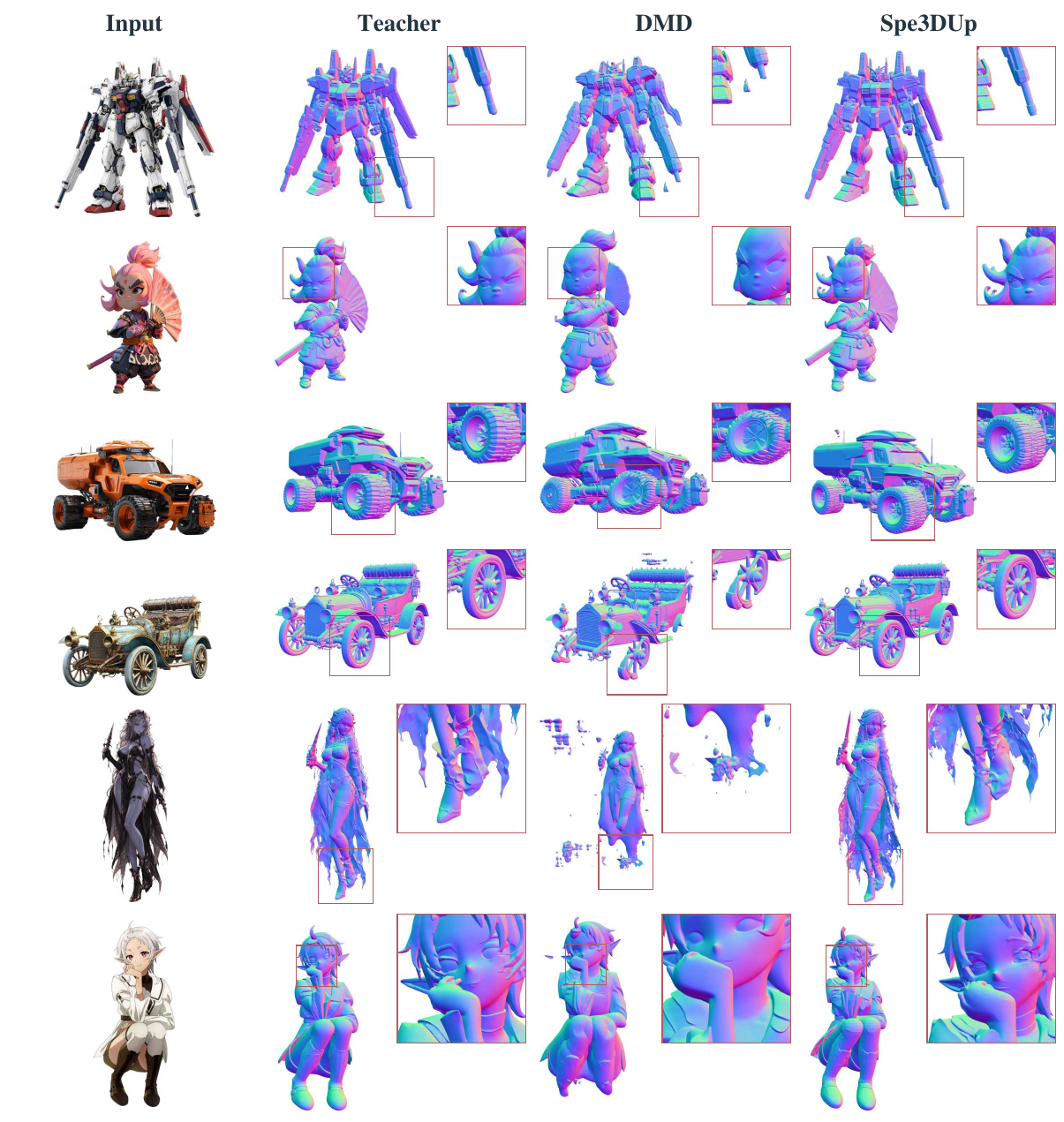}
\caption{\textbf{Qualitative mesh comparison.}
Normal renderings of final meshes and close-ups of the boxed regions.}
\label{fig:qualitative}
\end{figure*}

\begin{figure*}[b]
\centering
\includegraphics[width=\linewidth]{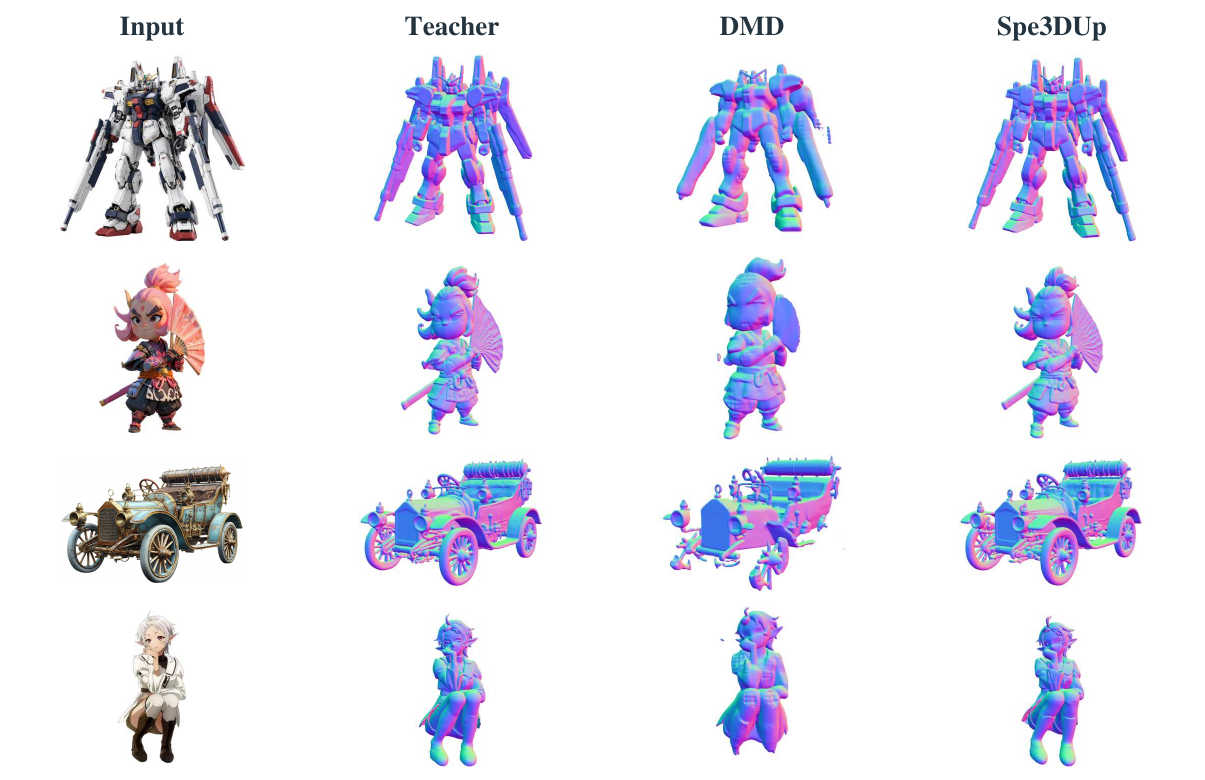}
\caption{\textbf{Coarse geometry comparison.}
Normal renderings before refinement.}
\label{fig:coarse_qualitative}
\end{figure*}

\paragraph{Coarse generation.}
\Cref{tab:main} shows that FTD preserves reconstruction quality close to the teacher in ten coarse-generation steps.
Its surface distance and F-score remain close to the teacher's, and its normal consistency is comparable, while DMD has larger reconstruction errors.
The coarse outputs in \cref{fig:coarse_qualitative} help explain this difference: the samurai's protruding facial features and the car's wheel structure are already weakened or missing in DMD before refinement begins.
FTD preserves these parts more completely, providing a more faithful starting geometry for subsequent detail recovery.

\paragraph{Cascaded generation.}
The advantage persists when each refiner receives its own generator's coarse output.
FTD achieves geometric reconstruction quality close to the full Teacher pipeline and exceeds DMD on all five reported measures (\cref{tab:main}).
The final meshes in \cref{fig:qualitative} show that the difference is not simply the amount of surface detail.
Broken weapon barrels, incomplete wheels, and fragmented lower limbs or garments change the identity-defining parts of the object, even when other regions remain detailed.
FTD retains more continuous and recognizable geometry in these regions.
Comparing the coarse and refined figures shows that some of the final discrepancies originate in coarse generation and remain after refinement; the shared-coarse experiment below examines the refiner separately.

\begin{figure*}[b]
\centering
\includegraphics[width=\linewidth]{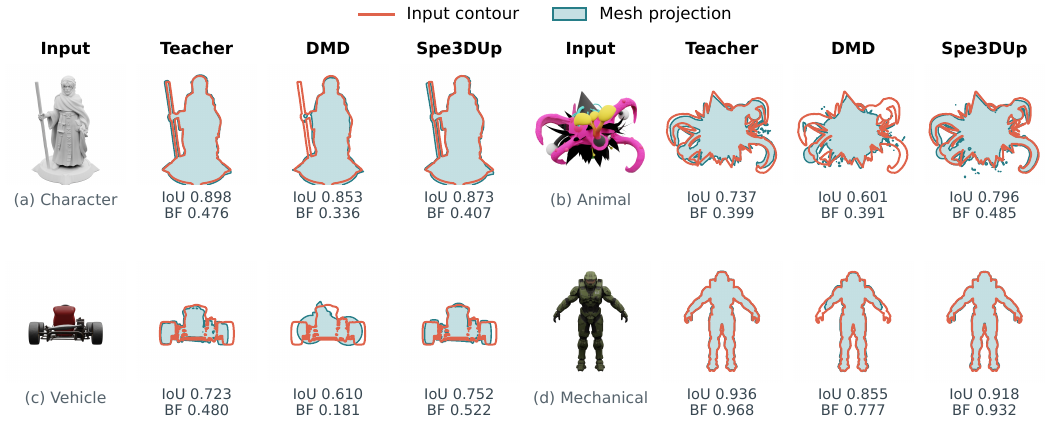}
\caption{\textbf{Input-silhouette comparison after pose and scale alignment.}
Red outlines indicate the input silhouette; teal regions show the mesh projection.}
\label{fig:rigid_input_alignment}
\end{figure*}

\paragraph{Input fidelity.}
FTD achieves higher silhouette IoU and ULIP-2 similarity than DMD for both coarse geometry and final meshes (\cref{tab:main}).
The silhouette comparison in \cref{fig:rigid_input_alignment} further shows that FTD better preserves the input object's overall proportions, main shape, and protruding parts, reducing the shape contraction and part displacement seen in DMD outputs.
Together with the part completeness visible in the qualitative results, these observations show that FTD more accurately preserves the reference object's specific structure, rather than merely generating a detailed mesh of the same category.

This advantage is consistent with FTD's paired trajectory supervision: the student matches teacher update endpoints under the same input conditions and starting states, directly learning the teacher's geometric response to each input.
This helps preserve correspondence to the reference image while compressing the sampling process.

\subsection{Distillation Analysis}
\label{sec:fixed_coarse}

We feed the same teacher-generated coarse geometry and reference image to the Teacher, DMD, and FTD refiners and compare their surface-reconstruction quality and input fidelity (\cref{tab:fixed_coarse}).

\fixedcoarserefinementtable

\paragraph{Effect of coarse geometry.}
With teacher coarse geometry, DMD's Chamfer distance ($\times10^3$) decreases from 29.1 in the complete cascade to 15.9, bringing its reconstruction quality substantially closer to the teacher's (\cref{tab:main,tab:fixed_coarse}).
This indicates that structural errors in coarse generation are an important source of the final quality gap.
Coarse geometry provides the spatial support for UltraShape refinement, so omissions and deviations in the initial structure affect subsequent surface recovery, consistent with the missing parts in \cref{fig:coarse_qualitative}.

\paragraph{Refinement quality.}
With the same coarse geometry, FTD still achieves lower surface distance, a higher F-score, and higher ULIP-2 similarity than DMD, indicating more accurate surface recovery while preserving correspondence to the reference image.
This result is consistent with FTD's paired trajectory supervision: the student matches teacher update endpoints under the same image, coarse geometry, and starting latent, learning the refinement process conditioned on the given input.
The appendix additionally reports cross-cascade results obtained by applying the DMD refiner to FTD coarse geometry.

\paragraph{Training budget and quality.}
The main comparisons use the primary DMD coarse and refinement checkpoints, with a combined training cost of approximately 570 A800 GPU-hours.
Across both stages, FTD uses approximately 1,300 A800 GPU-hours, including guidance distillation and interval compression.
For the budget study, we resume the primary DMD runs and continue training to a cumulative 1,400 GPU-hours across both stages, slightly exceeding FTD's recorded budget.
On the same test set, the extended checkpoints increase DMD's complete-cascade Chamfer distance ($\times10^3$) from 29.1 to 46.9, whereas its fixed-teacher-coarse refinement distance changes only from 15.9 to 15.8.
The coarse generator also deteriorates, indicating different responses to continued training across the two stages.
For this training recipe, increasing the budget does not close the complete-cascade quality gap.
Detailed training costs and checkpoint quality curves are provided in the appendix.

\subsection{End-to-end efficiency}
\label{sec:stage_ablation}
\label{sec:remesh_exp}

\Cref{fig:latency_breakdown} shows how the computational bottleneck shifts after sampling compression.
Coarse and refinement sampling become $\coarseDitSpeedup\times$ and $\refineDitSpeedup\times$ faster, respectively, but retaining the original decoding and serial mesh processing yields only a $2.7\times$ end-to-end speedup, with geometry recovery and processing now accounting for most of the latency.
Accelerating decoding and mesh simplification raises the complete-pipeline speedup to $\cumulativeSpeedupRounded\times$ relative to the original teacher.
With accelerated geometry recovery applied to both teacher and student, FTD provides a $5.9\times$ end-to-end speedup from sampling distillation.

\subsection{Geometry-recovery acceleration}
\label{sec:ablations}

\paragraph{Decoding efficiency and geometric quality.}
Accelerated decoding reduces geometry-recovery cost while keeping reconstruction error, F-scores, and normal consistency close to those of the original decoder (\cref{fig:latency_breakdown,tab:refinement_decode_gt}).

\refinementdecodegttable

\Cref{tab:decoder_four_path} separates the effects of the decoding optimizations: operator optimization reduces field-evaluation time from 25.6 to 14.1 seconds, and field reuse further reduces it to 11.1 seconds while increasing surface deviation from the reference mesh.
Critical-region recomputation takes 12.3 seconds, adding only 1.2 seconds over field reuse while reducing mean surface deviation by approximately 77\%.
These results show that field reuse provides most of the query savings, while additional evaluations in critical regions effectively correct the geometric errors it introduces.
Further quality and topology analyses are provided in the appendix.

\decoderprofilestable

\paragraph{Parallel mesh processing.}
\Cref{fig:latency_breakdown}(b) shows that using 16 workers reduces mean mesh-simplification time from 53.2 to 7.8 seconds, a $6.8\times$ speedup.
The parallel and serial outputs have a mean normal consistency of 0.973 and small surface distances, indicating that regional parallelism substantially reduces simplification time while retaining geometry close to the serial output.
Further latency and geometric-deviation results are provided in the appendix.

\begin{samepage}
\section{Conclusion}

Spe3DUp accelerates coarse-to-fine 3D generation by folding guidance and matching teacher interval endpoints.
Combined with efficient decoding and parallel QEM simplification, it achieves $\ditSpeedup\times$ DiT and $\systemSpeedup\times$ end-to-end acceleration with matched decoding and postprocessing.
Coarse, fixed-coarse refinement, and complete-cascade evaluations show that FTD retains geometric quality and input correspondence close to the teacher.
Cross-resolution field reuse and critical-region recomputation reduce reconstruction cost, while parallel simplification accelerates delivery of the final mesh.
\par
\end{samepage}

%% file: sections/algorithm.tex
% High-level overview; the appendix provides the full correction procedure.
\begin{algorithm}[!t]
\caption{Refinement field reuse and recomputation}
\label{alg:refinement_decode}
\begin{algorithmic}[1]
\algtext*{EndFor}
\algtext*{EndIf}
\Require Decoder $D$, field $P$, queries $Q$, batch map $\mathcal B$
\Statex $L_c,L_s$: iteration budgets
\Statex $\mathit{seen}$: queried points; $\mathit{done}$: recomputed batches
\Ensure High-resolution mesh
\State $F\gets\Call{Interpolate}{P,Q}$
\State $\mathit{seen}\gets\Call{Boundary}{P,Q}$; $\mathit{done}\gets\varnothing$
\State $F[\mathit{seen}]\gets D(\mathit{seen})$
\For{$k=1,\ldots,L_c$}
    \State $B\gets\mathcal B(\Call{Critical}{F})\setminus\mathit{done}$
    \State \textbf{if} $B=\varnothing$ \textbf{then break}
    \ForAll{$b\in B$}
        \State $F[b]\gets D(b)$; $\mathit{seen}\gets\mathit{seen}\cup b$
    \EndFor
    \State $\mathit{done}\gets\mathit{done}\cup B$
\EndFor
\If{$\mathcal B(\Call{Critical}{F})\setminus\mathit{done}\ne\varnothing$}
    \ForAll{$b\in\mathcal B(Q)\setminus\mathit{done}$}
        \State $F[b]\gets D(b)$; $\mathit{seen}\gets\mathit{seen}\cup b$
    \EndFor
\EndIf
\For{$k=1,\ldots,L_s$}
    \State $S\gets\Call{Surface}{F}\setminus\mathit{seen}$
    \State \textbf{if} $S=\varnothing$ \textbf{then break}
    \State $F[S]\gets D(S)$; $\mathit{seen}\gets\mathit{seen}\cup S$
\EndFor
\If{$\Call{Surface}{F}\setminus\mathit{seen}\ne\varnothing$}
    \State $F[Q\setminus\mathit{seen}]\gets D(Q\setminus\mathit{seen})$
\EndIf
\State \Return $\Call{MarchingCubes}{F,0}$
\end{algorithmic}
\end{algorithm}